\documentclass{article}

\usepackage[preprint]{corl_2026} 
\usepackage{amsfonts}
\usepackage{amssymb,mathtools}
\usepackage{algpseudocode}
\usepackage{algorithm}
\usepackage{nomencl}
\usepackage{tabularx}
\usepackage{amsmath}
\usepackage{subcaption}
\usepackage[numbers]{natbib}
\usepackage{enumitem}
\usepackage{cleveref}
\usepackage{siunitx}      
\usepackage{makecell}    
\usepackage{adjustbox}

\title{LAEI: Layered Autonomous Edge Intelligence Framework for Robust UAV Swarm Operations}

\author{ Changmin Park\thanks{These first co-authors contributed equally to this work.}\\ School of Electrical Engineering\\ Korea University\\ Seoul 02841, Korea\\ \texttt{minpark0120@korea.ac.kr} \And Wooyong Jung\footnotemark[1]\\ School of Electrical Engineering\\ Korea University\\ Seoul 02841, Korea\\ \texttt{jy17347@korea.ac.kr} \And Hwangnam Kim\thanks{Corresponding author.}\\ School of Electrical Engineering\\ Korea University\\ Seoul 02841, Korea\\ \texttt{hnkim@korea.ac.kr} }

\begin{document}
\maketitle


\begin{abstract}
Autonomous UAV swarms require scalable coordination mechanisms that maintain mission performance under limited communication, environmental uncertainty, and component failures. Centralized approaches provide global coordination but suffer from communication bottlenecks and single-node vulnerabilities, whereas fully decentralized methods often lack mission-level consistency. This paper presents Layered Autonomous Edge Intelligence (LAEI), a UAV-swarm framework that combines onboard learned policies with lightweight mission-level supervision. Each UAV performs local perception, obstacle avoidance, and action selection onboard, while the supervisory layer provides adaptive goal reassignment, fault-aware recovery, and context-dependent policy guidance without directly controlling low-level actions. LAEI further incorporates recovery strategies, including dynamic reassociation, backup supervisory support, and fallback local autonomy, to maintain mission continuity under representative failure scenarios. We evaluate LAEI in simulated UAV-swarm scenarios using mission completion time, collision rate, and coverage efficiency. The results show that LAEI reduces mission completion time and improves operational efficiency while maintaining collision-aware distributed UAV-level decision-making.

\end{abstract}

\keywords{UAV Swarm, Edge AI, Multi-Agent, Autonomous Robotics, Mission-Level Supervision} 


\section{Introduction}
	
The use of Unmanned Aerial Vehicles (UAVs) has expanded across surveillance, delivery, disaster response, and environmental monitoring \cite{uavApplications}. As mission requirements become more complex, UAV systems are increasingly shifting from single-agent operation to collaborative swarm operation, where multiple UAVs cooperate to improve coverage, robustness, and efficiency \cite{park}. However, scalable coordination remains challenging. Centralized swarm-control approaches provide global coordination but are vulnerable to communication bottlenecks and single-node failures \cite{centralizedControl, communicationChallenges}, while fully decentralized methods improve local autonomy but often struggle to maintain mission-level consistency under limited communication, partial observability, or agent failures \cite{distributedControl}.

Recent advances in reinforcement learning and edge intelligence enable UAVs to process local observations and select actions onboard under computational and communication constraints \cite{traditionalControlLimitations, aiAdaptability}. Nevertheless, onboard autonomy alone is insufficient for maintaining coherent swarm-level behavior when mission objectives change or failures occur. This motivates a layered coordination architecture that preserves UAV-level autonomy while providing lightweight mission-level guidance.

To address these challenges, this paper proposes Layered Autonomous Edge Intelligence (LAEI), a UAV-swarm framework that combines onboard edge-AI policies with a lightweight supervisory layer. In LAEI, each UAV performs local perception, obstacle avoidance, and action selection onboard, while the supervisory layer provides adaptive goal reassignment, fault-aware recovery signals, and context-dependent policy recommendations without directly controlling low-level UAV actions. This separation allows LAEI to balance distributed autonomy with mission-level supervision.

LAEI also incorporates fault-tolerant recovery strategies, including dynamic reassociation, backup supervisory support, and fallback local autonomy. These mechanisms reduce dependency on a single mission-level supervisor and allow UAVs to maintain safe degraded operation until connectivity is restored or reassignment becomes available.

The main contributions of this paper are summarized as follows:
\begin{itemize}[noitemsep, topsep=0pt]
\item We propose LAEI, a layered UAV-swarm framework that combines onboard edge-AI policies with mission-level supervision while preserving distributed decision-making.
\item We design supervisory coordination mechanisms for adaptive goal reassignment, context-dependent policy recommendation, and fault-aware recovery without direct action control.
\item We evaluate LAEI in simulated UAV-swarm scenarios and edge-device inference tests, demonstrating improvements in mission time, collision rate, and coverage efficiency.
\end{itemize}


\section{RELATED WORK}

UAV swarm coordination has been widely studied through centralized, decentralized, and hierarchical control approaches \cite{zhou}. Centralized or hierarchical control systems can provide global mission awareness and coordinated task allocation, but they often suffer from scalability limitations, communication bottlenecks, and vulnerability to central-node failures \cite{hierarchicalControl,HCA}. In contrast, decentralized and distributed methods allow UAVs to make decisions based on local observations and neighbor interactions, improving autonomy and flexibility \cite{DFA}. However, purely decentralized control can make mission-level consistency difficult to guarantee when local decisions must remain aligned with global objectives \cite{DFA_2}.

Learning-based approaches, including reinforcement learning and multi-agent reinforcement learning, have also been investigated for UAV swarm coordination \cite{jung2024enhancing, eke}. These methods enable agents to learn adaptive behaviors for coverage, exploration, path planning, and collision avoidance \cite{han}. Hierarchical multi-agent learning further decomposes decision-making into high-level and low-level policies, improving coordination in complex tasks \cite{hier}. However, when the high-level policy directly determines subgoals or agent-level behavior, the system may still depend heavily on the availability and reliability of the high-level module \cite{limits}.

Recent UAV systems increasingly adopt edge AI to support onboard autonomy. Unlike cloud-based approaches, edge intelligence enables local inference and decision-making with reduced latency and communication dependency \cite{re2}. Lightweight neural networks have been applied to onboard object detection and tracking on embedded processors \cite{re3}, and distributed or federated learning methods have been explored for UAV networks \cite{re5,re6,re8}. Nevertheless, edge intelligence alone does not fully address swarm-level coordination, particularly when UAVs fail, communication links degrade, or mission objectives change \cite{re7}.

LAEI is positioned between centralized and distributed approaches. Each UAV performs onboard perception, obstacle avoidance, and action selection through its edge-AI policy, while a lightweight supervisory layer provides mission-level guidance, adaptive goal reassignment, fault-aware recovery signals, and context-dependent policy recommendations. Unlike centralized or hierarchical controllers that directly determine low-level actions or subgoals, LAEI preserves UAV-level autonomy while supporting swarm-level consistency and degraded operation under failures.

\section{System Design}
This section presents the proposed LAEI framework. LAEI consists of mission UAVs equipped with onboard edge-AI policies and a lightweight mission-level supervisor. The key design principle is to separate local action selection from mission-level supervision: each UAV selects low-level actions onboard, while the mission-level supervisor provides high-level guidance for reassignment, parameter adaptation, and recovery.

\subsection{System Overview}
The proposed system consists of two main entities: mission UAVs and a mission-level supervisor. Mission UAVs collect local sensor data, exchange limited state information with nearby UAVs, and execute tasks through onboard edge-AI modules. Each UAV independently performs perception, obstacle avoidance, and action selection, enabling distributed swarm behavior without continuous low-level commands from the mission-level supervisor.

The mission-level supervisor provides lightweight mission-level supervision rather than direct action control. It updates target assignments, recommends policy-conditioning parameters, and provides recovery signals when mission objectives, UAV states, or failure conditions change. These signals are incorporated into the onboard policies while preserving UAV-level autonomy.

\begin{figure*}[t]
\centering
\includegraphics[width=0.75\linewidth]{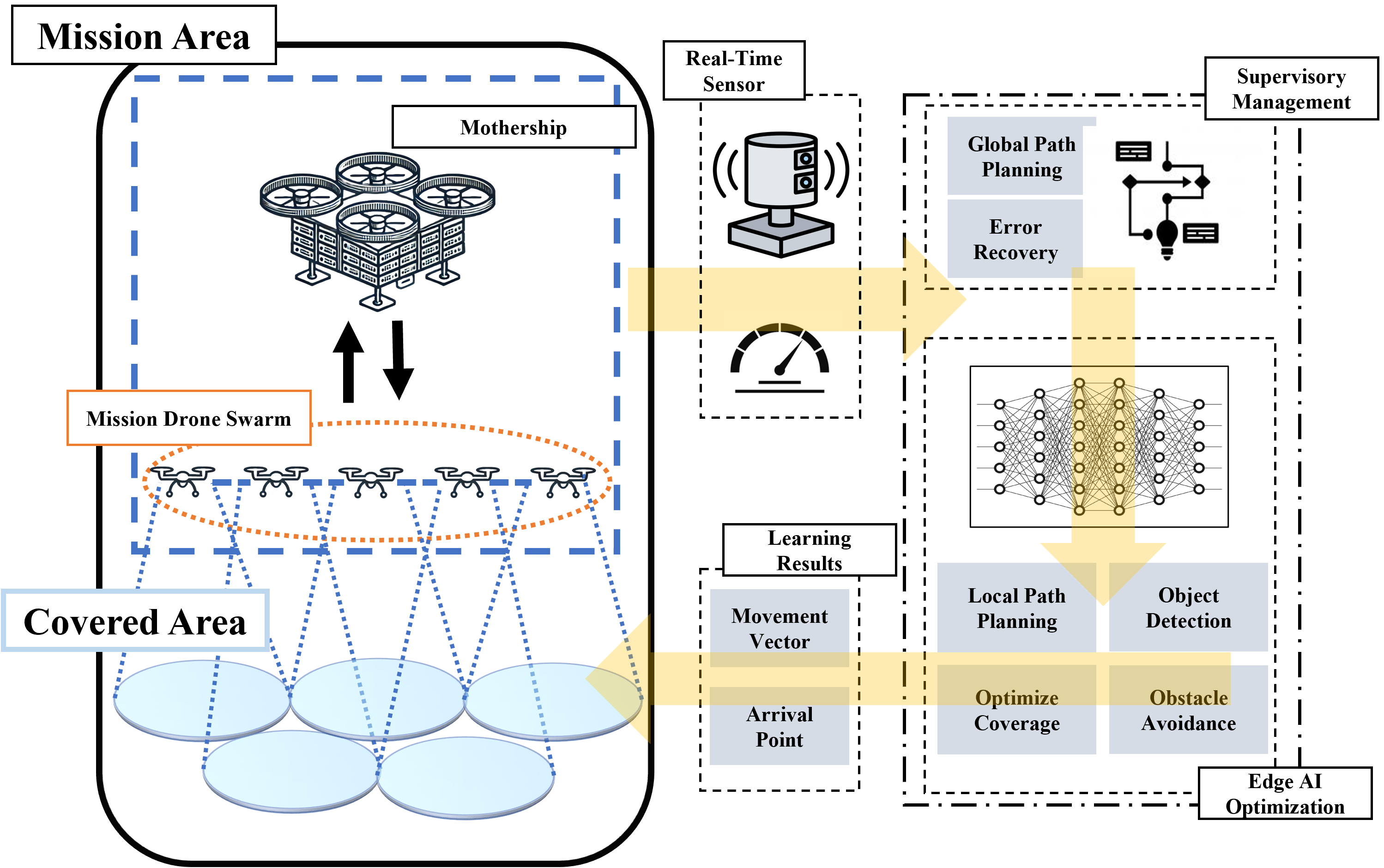}
\caption{System-level architecture of the proposed LAEI framework.}
\label{fig:system_overview}
\end{figure*}

As shown in Fig.~\ref{fig:system_overview}, LAEI combines onboard edge intelligence with mission-level supervision. During normal operation, UAVs react to local observations using their onboard policies. During mission updates or abnormal conditions, the mission-level supervisor provides reassignment and recovery information to maintain swarm-level consistency.

\subsection{Mission-Level Supervisory Coordination}
\label{subsec:supervisory_coordination}

The mission-level supervisory coordination module is implemented on the mothership, whereas each UAV performs local perception, collision avoidance, decision-making, and action execution through its onboard edge-AI module.

\subsubsection{Adaptive Goal Reassignment}

Adaptive goal reassignment is performed by the mothership to maintain mission efficiency as UAV positions and mission conditions change. Instead of solving a costly global assignment problem at every update, LAEI uses a lightweight pairwise exchange heuristic. Let $\pi$ denote the current assignment of UAVs to destinations. For each pair of UAVs $(i,j)$ with positions $p_i,p_j$ and assigned destinations $d_{\pi(i)},d_{\pi(j)}$, the current and swapped costs are defined as
\begin{equation}
C_{\text{cur}} = |p_i-d_{\pi(i)}| + |p_j-d_{\pi(j)}|, \quad
C_{\text{swap}} = |p_i-d_{\pi(j)}| + |p_j-d_{\pi(i)}|.
\end{equation}
A swap is accepted when
\begin{equation}
C_{\text{swap}}+\epsilon < C_{\text{cur}},
\end{equation}
where $\epsilon>0$ prevents oscillatory swaps. The updated assignment is transmitted to UAVs as revised mission guidance. The UAVs do not receive direct motion commands from the mothership; instead, each UAV uses its assigned destination as part of its local decision-making process.

\subsubsection{Context-Dependent Parameter Recommendation}

In addition to goal reassignment, the mothership recommends policy-conditioning parameters according to the current mission context:
\begin{equation}
\eta_i = [\alpha_{\text{safe}}, \alpha_{\text{avoid}}, v_{\max}, \tau_{\text{comm}}, \tau_{\text{reassign}}, \rho_{\text{recover}}],
\end{equation}
where the elements represent the safety margin, collision-avoidance weight, velocity bound, communication threshold, reassignment threshold, and recovery priority, respectively. These parameters are updated using lightweight rule-based decisions based on mission context, such as communication quality, local congestion, task progress, and detected failure conditions.

The recommended parameter vector $\eta_i$ is delivered to the corresponding UAV and used as a conditioning input for the onboard edge-AI policy. Thus, the mothership does not directly determine the action $a_i$ of UAV $i$. Instead, the onboard policy selects the action according to
\begin{equation}
a_i \sim \pi_\theta(a_i \mid o_i, d_{\pi(i)}, \eta_i),
\end{equation}
where $o_i$ is the local observation, $d_{\pi(i)}$ is the assigned destination, and $\eta_i$ is the mission-level conditioning vector. This design allows the mothership to influence UAV behavior at the mission level while preserving distributed onboard autonomy.

\subsubsection{Fault-Aware Recovery}

Fault-aware recovery is activated when the mothership detects UAV failure from status updates, missing heartbeats, or abnormal communication states. When a UAV $u_f$ fails, its assigned task is redistributed among the remaining UAVs $\mathcal{U}'=\mathcal{U}\setminus\{u_f\}$. To preserve coverage, the mothership samples candidate points around the original goal region and applies k-means clustering with $|\mathcal{U}'|$ clusters to generate updated virtual goals $\mathcal{D}^*$. These goals are refined by local stochastic perturbations accepted only when coverage improves.

Coverage is measured as
\begin{equation}
\text{Coverage}(P,\tilde{\mathcal{D}},r)=
\frac{
\left|\left\{p\in P \mid \min_{d\in\tilde{\mathcal{D}}}|p-d|\leq r\right\}\right|
}{
|P|
},
\end{equation}
where $P$ is the candidate point set and $r$ is the sensing radius. The operational UAVs are then reassigned to the updated destinations by minimizing total travel distance:
\begin{equation}
\pi^*=\arg\min_{\pi\in S_{|\mathcal{U}'|}}
\sum_{i\in\mathcal{U}'} \left|p_i-d_{\pi(i)}^*\right|.
\end{equation}
The mothership also increases the recovery priority $\rho_{\text{recover}}$ and may adjust the velocity bound, safety margin, or reassignment threshold to support stable degraded operation. The remaining UAVs then continue the mission using the updated assignment and policy-conditioning parameters.

Algorithm~\ref{alg:supervisory} summarizes the overall mothership-based supervisory coordination procedure. The algorithm clarifies how the three supervisory functions are integrated into a single mission-level loop. During normal operation, adaptive goal reassignment and context-dependent parameter recommendation are executed together. When a UAV failure is detected, the recovery process updates virtual goals, redistributes tasks, and adjusts recovery-related parameters. If the mothership becomes unavailable, UAVs continue operating in fallback local autonomy using the last valid mission guidance until mothership connectivity is restored.

\begin{algorithm}[t]
\caption{Mothership-Based Supervisory Coordination in LAEI}
\label{alg:supervisory}
\begin{algorithmic}[1]
\State \textbf{Input:} UAV observations ${o_i}$, positions ${p_i}$, goals ${d_j}$, assignment $\pi$, status ${s_i}$, communication quality ${q_i}$
\State \textbf{Output:} Updated assignment $\pi$ and parameters ${\eta_i}$
\State The mothership collects compact mission-state updates from UAVs
\If{UAVs do not receive updated guidance from the mothership}
\State UAVs use fallback local autonomy with the last valid $\pi$ and ${\eta_i}$
\ElsIf{UAV failure is detected}
\State Define operational UAV set $\mathcal{U}'$ and update virtual goals $\mathcal{D}^*$
\State Reassign goals and update recovery parameters ${\eta_i}$
\Else
\State Refine goal assignment $\pi$ using pairwise exchange
\State Recommend context-dependent parameters ${\eta_i}$
\EndIf
\State Transmit $\pi$ and ${\eta_i}$ to UAVs as mission-level guidance
\State Each UAV selects onboard action using $\pi_\theta(a_i \mid o_i,d_{\pi(i)},\eta_i)$
\end{algorithmic}
\end{algorithm}

\subsection{Onboard Edge-AI Policy}

Each mission UAV contains an onboard edge-AI execution module that autonomously converts mission-level guidance into low-level actions. This module is responsible for local perception, state encoding, goal-conditioned navigation, collision-aware motion selection, and task execution. The onboard edge-AI module directly interacts with the local environment and selects actions using only local observations and lightweight supervisory information.

The observation of UAV $i$ is represented as
\begin{equation}
o_i=[s_{\text{task}},s_{\text{abs}},\phi_{\text{neighbor}},\eta_i],
\end{equation}
where $s_{\text{task}}$ includes goal-related information, $s_{\text{abs}}$ includes the UAV's kinematic state, $\phi_{\text{neighbor}}$ includes neighboring-agent and obstacle information, and $\eta_i$ denotes the policy-conditioning parameters provided by the mothership. Based on this observation, the onboard policy selects an action as
\begin{equation}
a_i \sim \pi_\theta(a_i \mid o_i,d_{\pi(i)}).
\end{equation}

To support safe local motion, collision-related interaction risks are encoded as part of the neighboring-agent representation. Specifically, we use a Collision Avoidance Vector $\vec{R}_{ij}$ from neighboring UAV $j$ to UAV $i$:
\begin{equation}
\vec{R}_{ij} =
\left(
\frac{\max(0,-(\vec{v}_j-\vec{v}_i)\cdot(\vec{p}_j-\vec{p}_i))}
{|\vec{p}_j-\vec{p}_i|^3+\epsilon}
\right)
\left(
\frac{\vec{p}_i-\vec{p}_j}{|\vec{p}_i-\vec{p}_j|}
\right).
\label{eq:cav}
\end{equation}
The total avoidance vector is computed as
\begin{equation}
\vec{R}_i=\sum_{j\in\mathcal{N}_i}\vec{R}_{ij}.
\end{equation}
This vector is not the sole function of the edge-AI module; rather, it provides an additional safety-aware feature that helps the onboard policy account for dynamic interaction risks during local navigation and action selection.

The edge-AI policy is trained using multi-agent PPO (MAPPO). The reward for UAV $i$ combines goal progress, goal completion, collision penalty, and safe-distance maintenance:
\begin{equation}
r_i=c_dD_i+c_gG_i+c_cC_i+c_sS_i,
\end{equation}
where $D_i$ is the distance-based reward, $G_i$ is the goal-reaching reward, $C_i$ is the collision penalty, and $S_i$ encourages safe separation from neighboring agents.

\begin{algorithm}[ht]
\caption{Training of Onboard Edge-AI Policy}
\label{alg:onboard_edge_ai}
\begin{algorithmic}[1]
\State \textbf{Input:} Number of episodes $N_{\text{episodes}}$
\State \textbf{Output:} Trained onboard policies ${\pi_\theta}$
\State Initialize policies ${\pi_\theta}$ and replay buffer $\mathcal{D}$
\For{episode $=1$ to $N_{\text{episodes}}$}
\State Reset environment and observe initial UAV states
\For{each step}
\State Encode local observations, goal information, neighboring-agent states, and supervisory parameters
\State Compute safety-aware interaction features such as $\vec{R}_i$
\State Select onboard actions using $\pi_\theta(a_i \mid o_i,d_{\pi(i)},\eta_i)$
\State Execute actions, observe rewards and next states, and store transitions in $\mathcal{D}$
\EndFor
\State Update policies and value functions using PPO mini-batches
\EndFor
\end{algorithmic}
\end{algorithm}

\section{Performance Evaluation}
We evaluate LAEI in simulation and edge-device-in-the-loop inference settings to assess mission-level supervision, failure recovery, and onboard edge-AI feasibility.

\subsection{Implementation Setup}
The control policies were evaluated in the VMAS simulation environment, configured as a continuous 2D space with 5 mission UAVs and 8 static obstacles. Each UAV was required to reach its assigned goal while avoiding other agents and obstacles. LAEI was compared with five baselines: A*, Information-Gain A*, ORCA, GWO, and PPO without supervisory guidance.

To validate onboard feasibility, we used an edge-device-in-the-loop setup in which the VMAS environment ran on a host PC while the trained edge-AI policy was executed on an NVIDIA Jetson Nano \cite{bettini2022vmas}. At each control step, the simulator transmitted the observation $o_i$ to the Jetson Nano, which computed the action $a_i$ and returned it to the simulator.

\subsection{Mission-Level Supervisory Management}
\begin{figure*}[t]
\centering
\begin{subfigure}{0.32\textwidth}
\centering
\includegraphics[width=\linewidth]{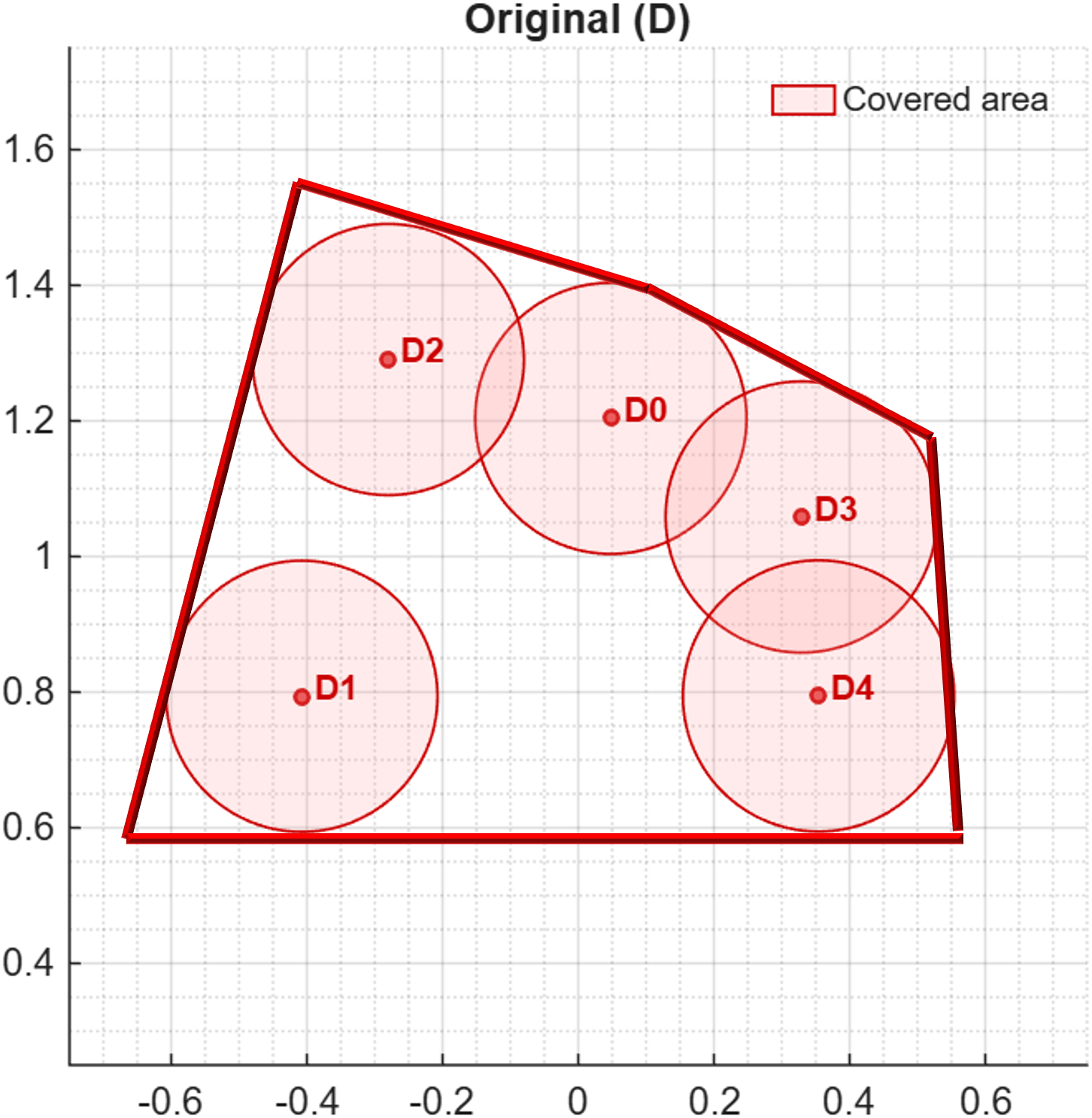}
\caption{Original ($D$)}
\label{fig:cov-a}
\end{subfigure}
\hfill
\begin{subfigure}{0.32\textwidth}
\centering
\includegraphics[width=\linewidth]{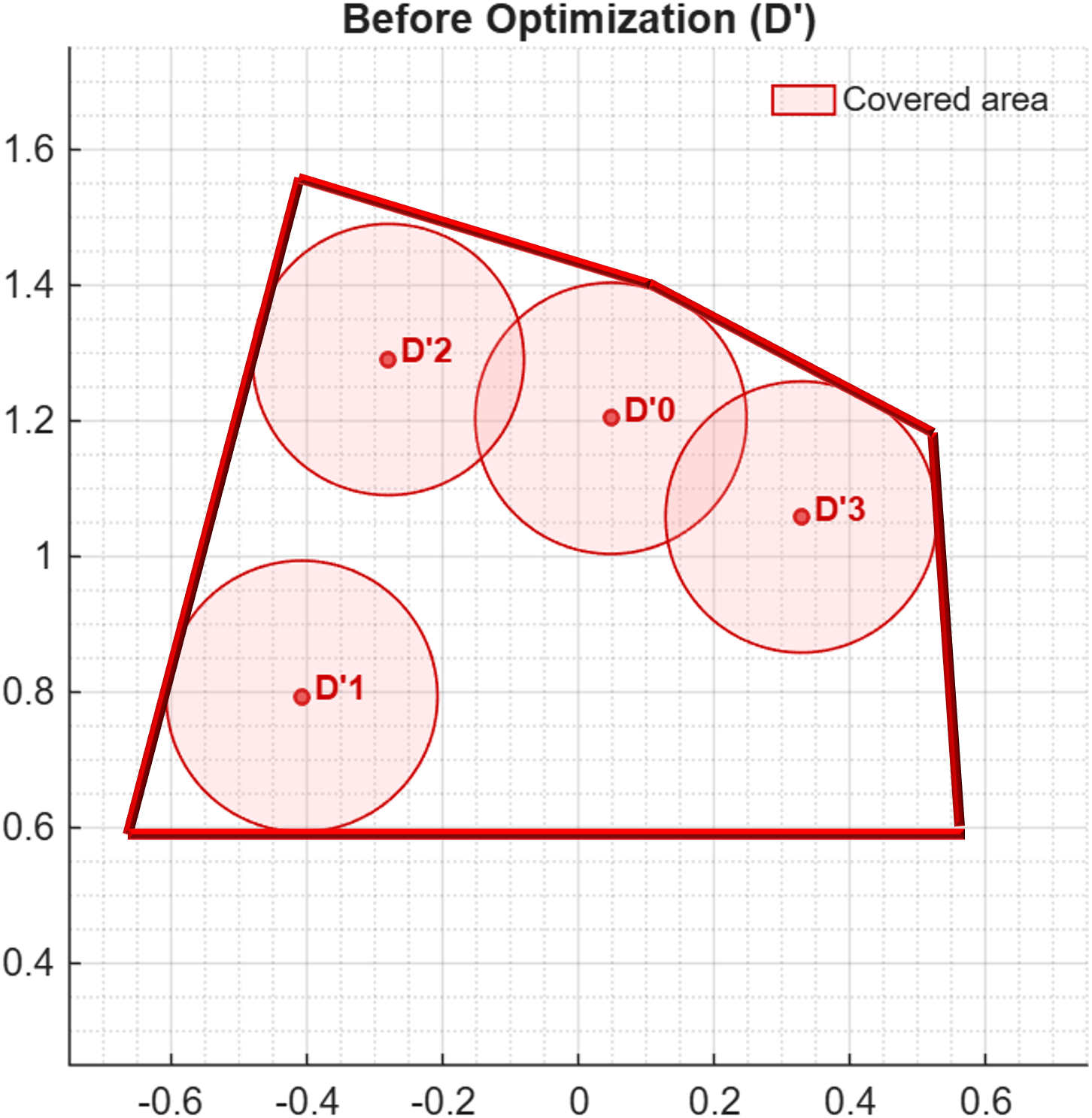}
\caption{Before ($D'$)}
\label{fig:cov-b}
\end{subfigure}
\hfill
\begin{subfigure}{0.32\textwidth}
\centering
\includegraphics[width=\linewidth]{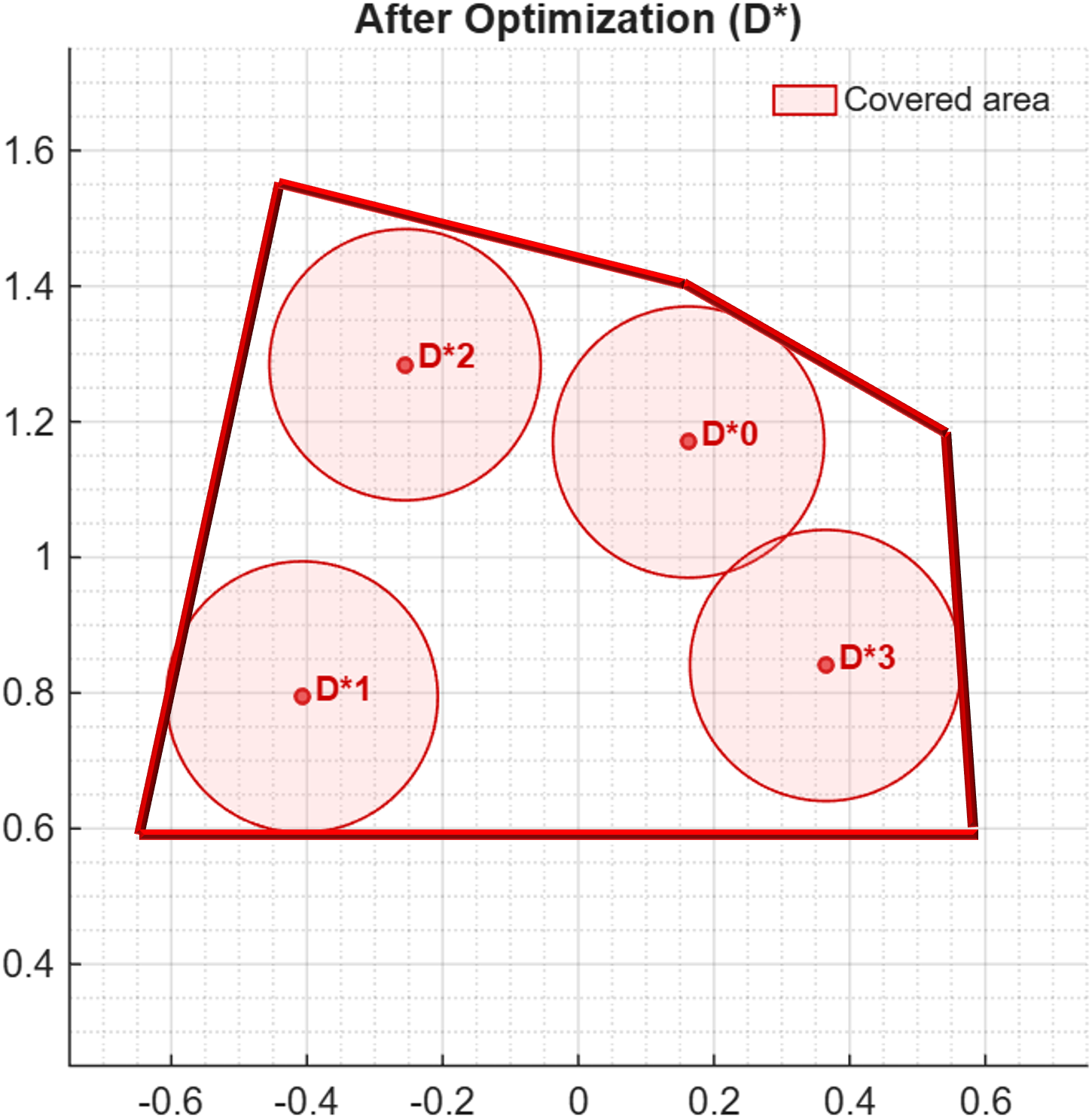}
\caption{After ($D^*$)}
\label{fig:cov-c}
\end{subfigure}
\caption{Coverage comparison before and after failure-aware goal optimization.}
\label{fig:coverage-three}
\end{figure*}

We first evaluated the effectiveness of mission-level supervisory management in adaptive goal assignment and failure recovery. In the adaptive goal assignment experiment, the mothership periodically reassigned targets to reduce total travel distance. A single target-swap operation reduced the cumulative distance from 4.9606 to 4.9217, corresponding to a 0.78\% improvement without global path replanning. For failure recovery, one UAV was removed from the swarm, and the remaining four goal positions were optimized to preserve coverage. As shown in Table~\ref{tab:coverage_res} and Fig.~\ref{fig:coverage-three}, the proposed optimization increased the coverage ratio from 83.12\% to 85.76\%, indicating improved coverage retention after UAV loss.

\begin{table}[t]
    \centering
    \caption{Coverage ratio before and after optimization.}
    \label{tab:coverage_res}
    \begin{tabular}{cc}
        \hline
        \textbf{Configuration} & \textbf{Coverage Ratio (\%)} \\
        \hline
        Before Optimization ($D'$) & 83.12 \\
        After Optimization ($D^*$) & 85.76 \\
        \hline
    \end{tabular}
\end{table}

\subsection{Edge-AI Performance}
\begin{figure*}[t]
\centering
\begin{subfigure}{0.48\textwidth}
\centering
\includegraphics[width=\linewidth]{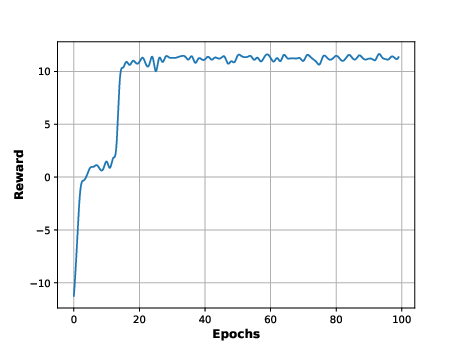}
\caption{Individual PPO}
\label{fig:rew_ppo}
\end{subfigure}
\hfill
\begin{subfigure}{0.48\textwidth}
\centering
\includegraphics[width=\linewidth]{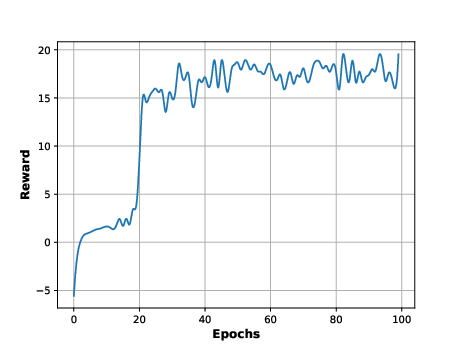}
\caption{LAEI}
\label{fig:rew_laei}
\end{subfigure}
\caption{Reward comparison between Individual PPO and LAEI.}
\label{fig:reward_comparison}
\end{figure*}

The reward curves in Fig.~\ref{fig:reward_comparison} show that both PPO and LAEI learn successfully, but LAEI converges to a higher reward value of approximately 18--19, whereas PPO plateaus around 11--12. This suggests that supervisory guidance helps stabilize learning and improve task-level performance.

We quantitatively evaluated each algorithm using collision count, completion time, coverage, and efficiency, where efficiency is computed as coverage divided by time. As summarized in Table~\ref{tab:comparison}, LAEI achieved zero collisions, the shortest completion time of 84 steps, and the highest efficiency of 0.034. Compared with PPO, LAEI reduced completion time by 16\% and improved efficiency by approximately 14\%, while maintaining zero collisions in the evaluated scenarios

\begin{table*}[b]
\centering
\caption{Performance comparison.}
\label{tab:comparison}
\resizebox{0.9\textwidth}{!}{%
\begin{tabular}{lcccccc}
\hline
\textbf{Metric} & \textbf{GWO} & \textbf{A$^\ast$} & \textbf{Information A$^\ast$} & \textbf{ORCA} & \textbf{PPO} & \textbf{LAEI} \\
\hline
Collision & 0 & 0 & 0 & 4 & 0 & 0 \\
Time & 140 & 121 & 107 & 116 & 100 & 84 \\
Coverage & 3.42 & 2.63 & 2.76 & 2.44 & 2.95 & 2.85 \\
Efficiency & 0.024 & 0.022 & 0.026 & 0.021 & 0.029 & 0.034 \\
\hline
\end{tabular}%
}
\end{table*}

\begin{figure*}[t]
\centering
\begin{subfigure}{0.32\textwidth}
\centering
\includegraphics[width=\linewidth,trim={5pt 5pt 5pt 5pt},clip]{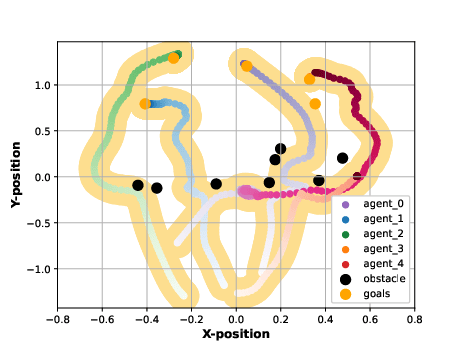}
\caption{GWO}
\label{fig:gwo}
\end{subfigure}
\hfill
\begin{subfigure}{0.32\textwidth}
\centering
\includegraphics[width=\linewidth,trim={5pt 5pt 5pt 5pt},clip]{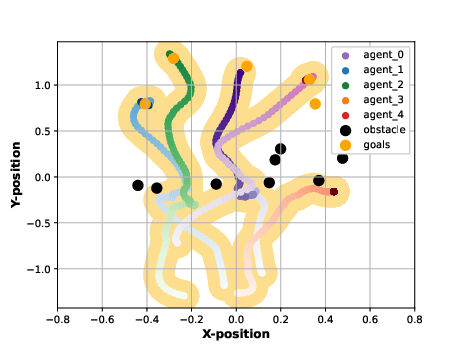}
\caption{A*}
\label{fig:a}
\end{subfigure}
\hfill
\begin{subfigure}{0.32\textwidth}
\centering
\includegraphics[width=\linewidth,trim={5pt 5pt 5pt 5pt},clip]{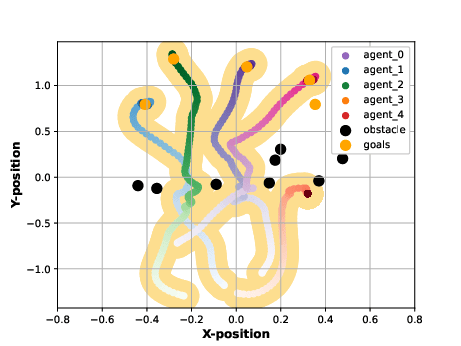}
\caption{Info-A*}
\label{fig:infoa}
\end{subfigure}

\vspace{-0.05cm}

\begin{subfigure}{0.32\textwidth}
    \centering
    \includegraphics[width=\linewidth,trim={5pt 5pt 5pt 5pt},clip]{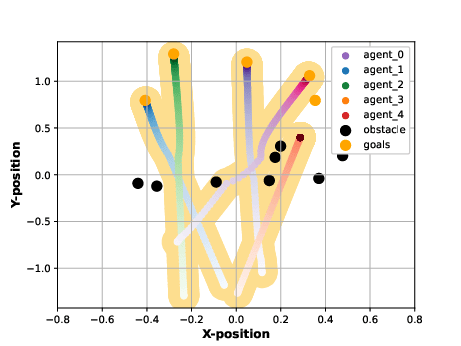}
    \caption{ORCA}
    \label{fig:orca}
\end{subfigure}
\hfill 
\begin{subfigure}{0.32\textwidth}
    \centering
    \includegraphics[width=\linewidth,trim={5pt 5pt 5pt 5pt},clip]{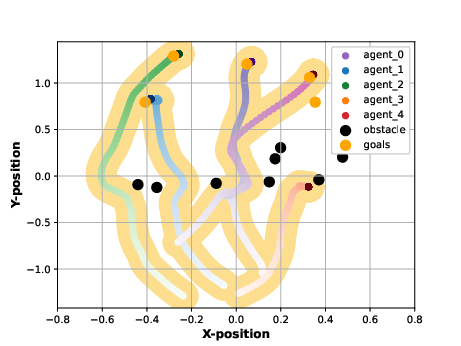}
    \caption{IPPO}
    \label{fig:non_swap}
\end{subfigure}
\hfill 
\begin{subfigure}{0.32\textwidth}
    \centering
    \includegraphics[width=\linewidth,trim={5pt 5pt 5pt 5pt},clip]{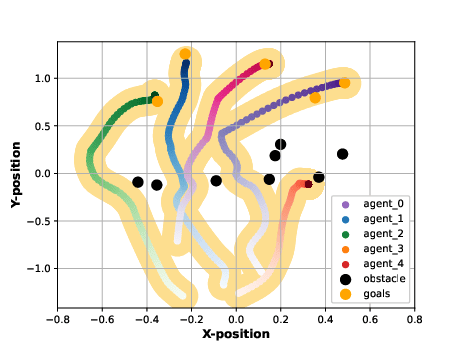}
    \caption{LAEI}
    \label{fig:laei}
\end{subfigure}

\caption{Trajectory comparison across baseline algorithms and LAEI.}
\label{fig:trajectories}
\end{figure*}

The trajectory results in Fig.~\ref{fig:trajectories} qualitatively support the quantitative results, showing that LAEI produces direct and smooth paths while avoiding obstacles and other agents.

\begin{figure*}[t]
\centering
\begin{subfigure}{0.48\textwidth}
\centering
\includegraphics[width=\linewidth]{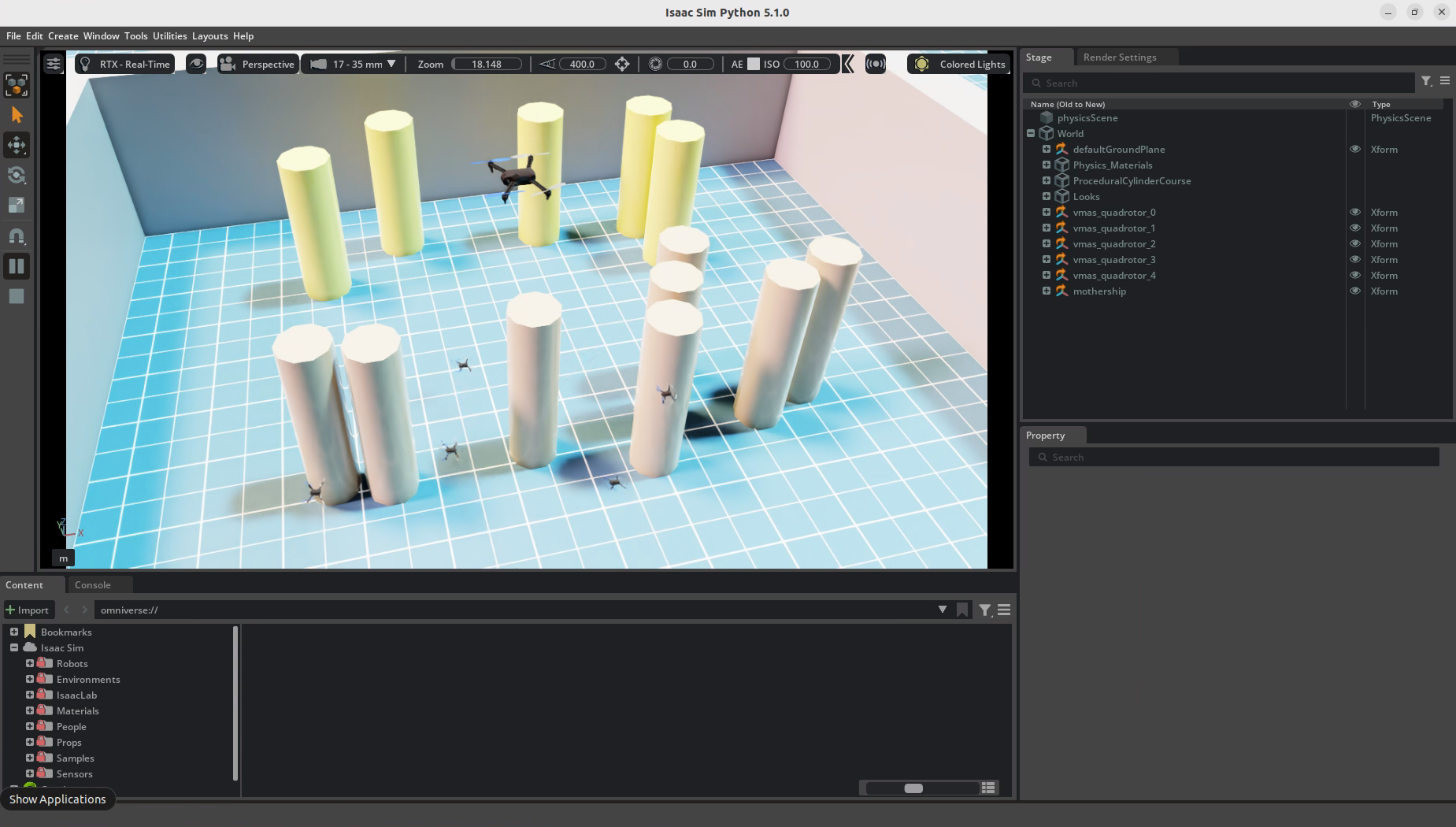}
\caption{Inference mission}
\label{fig:Inference}
\end{subfigure}
\hfill
\begin{subfigure}{0.48\textwidth}
\centering
\includegraphics[width=\linewidth]{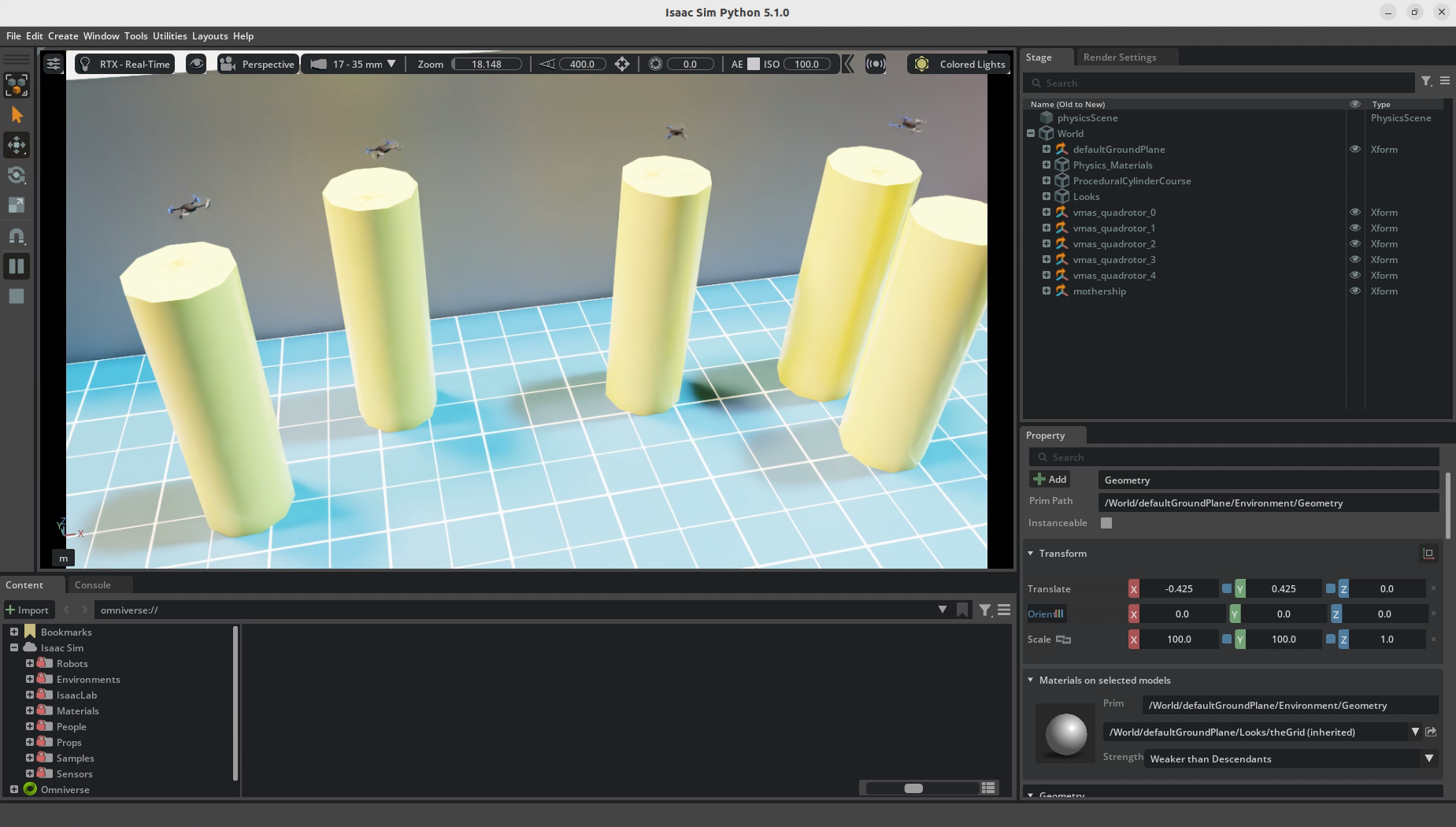}
\caption{Complete mission}
\label{fig:completion}
\end{subfigure}
\caption{Swarm Drone performs and completes mission in Isaac Sim.}
\label{fig:IsaacSim}
\end{figure*}

\subsection{Implementation and Validation}

Beyond the VMAS-based evaluation and hardware-in-the-loop inference test described in Section 4.1, we further validated the VMAS-trained policy in a higher-fidelity Isaac Sim environment with PX4-based UAV models. The policy was integrated into the Isaac Sim validation pipeline and applied to multiple PX4-based mission UAVs during mission execution. As shown in Fig.~\ref{fig:Inference}, the large UAV in the upper region represents the mothership, while the UAVs in the lower mission area represent the mission UAVs. This experiment examined whether the learned edge-AI policy could be transferred to a realistic multi-UAV simulation environment without additional retraining.

The validation was conducted on a workstation equipped with two NVIDIA RTX A6000 GPUs and an AMD Ryzen Threadripper PRO 5955WX CPU. Since the policy was trained in the VMAS coordinate space, we applied a coordinate-scale transformation to map policy-level positions and actions to the Isaac Sim mission space while preserving the relative mission geometry learned in VMAS.

We also implemented a degraded-operation scenario in which one mission UAV was assumed to be damaged during the mission. As shown in Fig.~\ref{fig:completion}, the remaining mission UAVs continued mission execution using the transferred policy and failure-aware mission configuration. The results confirm that the VMAS-trained edge-AI policy can be transferred to a higher-fidelity Isaac Sim--PX4 multi-UAV simulation environment.

\section{Conclusions}

This paper proposed LAEI, a layered UAV swarm framework that integrates onboard edge-AI policies with lightweight mission-level supervision. By separating local action selection from supervisory coordination, LAEI enables UAVs to operate autonomously while supporting adaptive goal assignment and fault-aware recovery. Simulation and edge-device-in-the-loop evaluations showed that LAEI improves mission efficiency and failure-aware recovery performance. In the evaluated scenarios, LAEI increased post-failure coverage from 83.12\% to 85.76\%, achieved zero collisions, reduced mission completion time to 84 steps, and obtained the highest efficiency score of 0.034 among the compared methods.

\newpage

\section{Limitations}

Although LAEI demonstrates improved coordination, efficiency, and fault-aware recovery in the evaluated scenarios, several limitations remain. First, the current evaluation is conducted mainly in a 2D simulation environment with a limited number of UAVs and static obstacles, which may not fully capture the complexity of real-world aerial missions. Second, while hardware-in-the-loop experiments with Jetson Nano validate the feasibility of onboard inference, the framework has not yet been deployed on physical UAV platforms. Third, communication degradation, sensing noise, environmental disturbances, and UAV failures are not yet fully modeled. Therefore, future work should extend LAEI to full 3D flight dynamics, larger swarm sizes, more realistic communication and sensing conditions, and real-world flight experiments.

\bibliography{example}
\end{document}